\documentclass[preprint,12pt,authoryear]{elsarticle}

\usepackage{amssymb}

\usepackage{makecell}

\usepackage{algorithm}
\usepackage{algorithmic}

\usepackage{multirow}

\usepackage{pifont}

\usepackage{graphicx}
\usepackage{amsmath}
\usepackage{amssymb}
\usepackage{booktabs}

\usepackage{setspace}
\usepackage{hyperref}
\usepackage[hyphenbreaks]{breakurl}

\usepackage{geometry}
\usepackage[capitalize]{cleveref}
\crefname{section}{Sec.}{Secs.}
\Crefname{section}{Section}{Sections}
\Crefname{table}{Table}{Tables}
\crefname{table}{Tab.}{Tabs.}

\begin{document}

\begin{frontmatter}



\title{Adversarial Infrared Geometry: Using Geometry to Perform Adversarial Attack against Infrared Pedestrian Detectors}

\author[label1]{Kalibinuer Tiliwalidi}
\ead{202011081727@std.uestc.edu.cn}

\author[label1]{Chengyin Hu}
\ead{cyhuuestc@gmail.com}

\author[label1]{Weiwen Shi}
\ead{weiwen_shi@foxmail.com}

\author[label1]{Ling Tian\corref{cor1}}
\ead{lingtian@uestc.edu.cn}

\affiliation[label1]{organization={University of Electronic Science and Technology of China},
            addressline={No. 2006, Xiyuan Avenue, Gaoxin District},
            city={Chengdu},
            postcode={611731},
            state={Sichuan},
            country={China}}


\cortext[cor1]{Corresponding author}
%

\begin{abstract}

Currently, infrared imaging technology enjoys widespread usage, with infrared object detection technology experiencing a surge in prominence. While previous studies have delved into physical attacks on infrared object detectors, the implementation of these techniques remains complex. For instance, some approaches entail the use of bulb boards or infrared QR suits as perturbations to execute attacks, which entail costly optimization and cumbersome deployment processes. Other methodologies involve the utilization of irregular aerogel as physical perturbations for infrared attacks, albeit at the expense of optimization expenses and perceptibility issues.
In this study, we propose a novel infrared physical attack termed Adversarial Infrared Geometry (\textbf{AdvIG}), which facilitates efficient black-box query attacks by modeling diverse geometric shapes (lines, triangles, ellipses) and optimizing their physical parameters using Particle Swarm Optimization (PSO). Extensive experiments are conducted to evaluate the effectiveness, stealthiness, and robustness of AdvIG. In digital attack experiments, line, triangle, and ellipse patterns achieve attack success rates of 93.1\%, 86.8\%, and 100.0\%, respectively, with average query times of 71.7, 113.1, and 2.57, respectively, thereby confirming the efficiency of AdvIG. Physical attack experiments are conducted to assess the attack success rate of AdvIG at different distances. On average, the line, triangle, and ellipse achieve attack success rates of 61.1\%, 61.2\%, and 96.2\%, respectively. Further experiments are conducted to comprehensively analyze AdvIG, including ablation experiments, transfer attack experiments, and adversarial defense mechanisms.
Given the superior performance of our method as a simple and efficient black-box adversarial attack in both digital and physical environments, we advocate for widespread attention to AdvIG.

\end{abstract}




\end{frontmatter}



\section{Introduction}

Deep neural networks (DNNs) have demonstrated remarkable achievements across various domains such as computer vision \cite{ref1}, natural language processing \cite{ref3, ref48}, and speech recognition \cite{ref2}, smart system \cite{ref49}, internet of things \cite{ref50, ref51, ref52}. Among these, computer vision stands out as a significant application area, focusing on the comprehension and analysis of images and videos. Despite their success, DNNs are susceptible to adversarial attacks \cite{ref5}, particularly in computer vision tasks.
Adversarial attacks \cite{ref6, AdvNB} represent a class of techniques aimed at exploiting vulnerabilities in DNNs. In such attacks, adversaries introduce subtle, intentional alterations to input data, aiming to manipulate the network into producing erroneous outputs. This manipulation can result in misleading outcomes, even when the model performs satisfactorily under normal circumstances. In the realm of computer vision, adversarial attacks primarily target images, where attackers apply minor modifications to induce incorrect predictions in image classification tasks.
Broadly speaking, digital adversarial attacks \cite{ref7,ref8} operate by introducing imperceptible perturbations to digital images, thereby challenging DNNs. In contrast, physical attacks \cite{ref9,ref10, AdvCL} diverge from digital methods. Here, attackers must physically modify target objects in the real world and subsequently capture images using cameras to input them into DNN models for adversarial exploitation. In the context of physical attacks, adversaries must navigate a trade-off between stealthiness and robustness. Essentially, smaller perturbations enhance stealthiness but compromise robustness, whereas larger perturbations offer increased robustness at the expense of concealment.

Object detectors \cite{ref4} play a crucial role in understanding images and videos with greater precision, serving as the foundation for numerous practical applications such as autonomous driving, video surveillance, and face recognition. Therefore, assessing the security of object detectors holds significant research importance.
Presently, most physical adversarial attacks against object detectors occur in visible light environments \cite{ref11,ref12}, involving the optimization of color textures and their placement as posters to deceive visible detectors (trained using visible light datasets) into misidentifying or failing to identify target objects. However, in infrared environments, target objects are typically presented in black and white patterns, rendering visible perturbations ineffective against infrared detectors (trained using infrared datasets). Infrared sensors solely detect the temperature of target objects. Consequently, physical attacks on infrared detectors \cite{ref13,ref14} typically involve selecting heating/insulation materials as physical perturbations, resulting in white and black perturbation patterns respectively under infrared sensors.
Among existing infrared physical attacks, the Bulb attack \cite{ref13} utilizes a bulb plate as a heating material, resulting in weak perturbation patterns and inadequate stealthiness. The QR attack \cite{ref14} employs aerogel to create an infrared invisibility cloak for multi-view infrared attacks, yet it also lacks true invisibility. Similarly, the AIP method \cite{ref15} utilizes irregular aerogel as thermal insulation material for infrared attacks but cannot perform multi-scale attacks and lacks stealthiness. These approaches typically deploy perturbations in front of or outside pedestrians, compromising stealthiness.
To address this issue, HCB \cite{ref16} proposes using a cold sticker as a cooling material placed inside clothing to enhance stealthiness. However, this method exhibits low optimization efficiency and weak robustness. Conversely, AdvCloth \cite{AdvCloth} employs an electric heating sheet as a physical perturbation deployed within the pedestrian's clothing to create infrared invisibility cloths. Nevertheless, the complex deployment process of this method poses practical challenges.
Table \ref{Table1} presents a comparison between existing methods and the proposed AdvIG.

\begin{table*} 
	\centering
 
    \setlength{\belowcaptionskip}{5pt}
    \caption{Performance comparison between existing infrared attack techniques and the proposed AdvIG.}
    \label{Table1}
	\begin{tabular}{ccccccc}

    \hline
    ~&Bulb attack&QR attack&AIP&AdvCloth&HCB&AdvIG \\
    \hline
    Stealthiness&\ding{55}&\ding{55}&\ding{55}&\ding{51}&\ding{51}&\ding{51}\\
    \hline
    Scenario&White-box&White-box&White-box&White-box&Black-box&Black-box\\
    \hline
    Efficiency&\ding{55}&\ding{55}&\ding{55}&\ding{55}&\ding{55}&\ding{51}\\
    \hline

    \end{tabular}
\vspace{0.3cm}
\end{table*}

Based on the above discussion, the current technical challenges in physical adversarial attacks for infrared detectors can be summarized as follows:
(1) Limited Perturbation Modes: Infrared perturbations are typically limited to black and white modes, restricting the diversity and granularity of perturbations that can be applied. This limitation hinders the ability to create more nuanced and effective adversarial samples.
(2) Efficient Optimization: Optimizing infrared adversarial samples within these limited perturbation modes poses a significant challenge. Efficiently optimizing perturbations to achieve desired adversarial effects while minimizing the perturbation magnitude is essential but challenging due to the constrained space for manipulation.

\begin{figure*}
\centering
\includegraphics[width=1\linewidth]{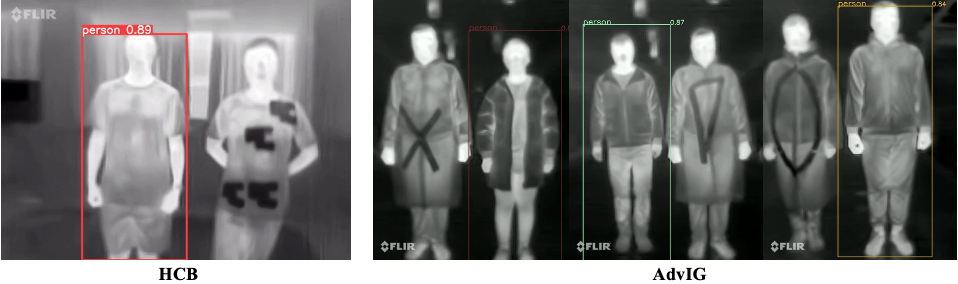} 
\caption{The existing black-box infrared physical attack comparison.}
\label{figure1}
\end{figure*}

To tackle the aforementioned challenges, we introduce a straightforward and effective approach for infrared black-box query attacks, named Adversarial Infrared Geometry (AdvIG). Illustrated in Figure \ref{figure1}, AdvIG adeptly executes physical adversarial attacks against infrared pedestrian detectors using basic geometric shapes such as lines, triangles, and ellipses. Our method initially optimizes the physical parameters of these geometric shapes using Particle Swarm Optimization (PSO) \cite{ref38} within a digital environment, efficiently identifying the most adversarial geometric perturbation patterns. Subsequently, leveraging these optimized physical parameters, we deploy cold patches as physical perturbations affixed inside pedestrian clothing within the physical environment. Ultimately, infrared sensors capture physical samples to execute infrared physical attacks.
Figure \ref{figure1} showcases physical samples generated by both the black-box infrared attack HCB \cite{ref16} and our proposed method. It is evident from the comparison that our approach produces perturbations that are more continuous and smooth, thereby enhancing the stealthiness of the infrared perturbation. Furthermore, our method utilizes affordable consumables such as cold stickers, keeping the total cost under \$5, which significantly simplifies the implementation process.
Our primary contributions can be summarized as follows:

\begin{itemize}
    \item We introduce Adversarial Infrared Geometry (AdvIG), a cost-effective and practical black-box infrared adversarial attack method. AdvIG is distinguished by its stealthiness, efficiency, and affordability, with a total cost of less than \$5. Its capability to perform black-box adversarial attacks further enhances its practicality.
    \item Extensive experiments are conducted to assess the effectiveness, stealthiness, and robustness of AdvIG. Results demonstrate that AdvIG achieves attack success rates exceeding 85\% and 50\% using lines, triangles, and ellipses in both digital and physical environments, affirming its effectiveness. Regarding stealthiness, our method strategically deploys perturbations inside pedestrian clothing, rendering them imperceptible to human observers without infrared sensors, while maintaining perturbation levels below baseline, confirming its stealthiness. Moreover, robustness tests reveal high attack success rates against various advanced infrared object detectors, validating its robustness.
    \item A series of comprehensive experiments have been conducted to thoroughly analyze AdvIG, encompassing ablation experiments, digital/physical transfer attack experiments, and adversarial defense experiments, among others. These experiments provide valuable insights into the performance of AdvIG. Additionally, we anticipate future research directions in the field of infrared physical attacks.
\end{itemize}

\section{Related works}

\subsection{Physical attacks in the visible light field}

\textbf{Patch-based physical attacks.} Patch-based physical attacks involve designing and generating adversarial perturbations using optimization methods. These perturbations are then applied to the target object or camera lens in the form of posters, stickers, etc., to conduct physical adversarial attacks on deep learning models. Common scenarios for patch-based physical adversarial attacks include face recognition, pedestrian detection, vehicle detection, and similar applications.
Sharif et al. introduced AdvGlass \cite{AdvGlass}, which attacks face recognition systems by deploying perturbations on eyeglass frames. Their approach utilizes an adversarial generative network to train a generator neural network, enhancing the adaptability of generated adversarial examples across different domains. AdvGlass exhibits robustness, concealment, and scalability, as demonstrated by experimental results.
Komkov et al. proposed AdvHat \cite{AdvHat} to target face recognition systems by attaching rectangular stickers printed by ordinary printers onto hats. They introduced a novel image texture transformation algorithm to simulate perturbation positions on hats, effectively confusing advanced face recognition models. AdvHat demonstrates superior portability in experimental evaluations.
Yin et al. presented AdvMakeup \cite{AdvMakeup}, a novel adversarial attack method that deceives face recognition systems using less suspicious makeup-based perturbations. Their approach employs a task-driven makeup generation method based on hybrid modules to synthesize invisible adversarial eye shadow in the orbital region of human faces. AdvMakeup incorporates a fine-grained meta-learning strategy to enhance attack transferability. Experimental results highlight the method's stealthiness in both digital and physical environments, making it deployable against advanced commercial systems.
Tan et al. introduced LAP \cite{LAP}, which aims to deceive both humans and object detectors simultaneously. Their approach employs a two-stage training policy framework to generate perturbation patterns that appear reasonable to the human eye while effectively fooling object detectors. LAP demonstrates significant adversarial and visual effects in experimental evaluations.
Hu et al. proposed NP \cite{NP} to enhance attack concealment by addressing perturbation visual rationality. Their approach leverages pre-trained generative adversarial networks to learn real-world images and generate natural adversarial patches under high adversarial conditions. Experimental results indicate NP's superior adversarial effects, with subjective investigations confirming the naturalness of generated adversarial examples compared to baselines.
To address multi-view adversarial attacks, Hu et al. introduced AdvTexture \cite{AdvTexture}, which generates adversarial textures with repeated structures and deploys them on clothing to conduct multi-view adversarial attacks against pedestrian object detectors. Experimental results demonstrate the effectiveness of clothing adorned with proposed adversarial textures in performing multi-view adversarial attacks.
Wang et al. proposed FCA \cite{FCA}, which renders non-planar camouflage textures on entire vehicle surfaces and introduces a transformation function to convert rendered camouflage vehicles into realistic scenes. Experimental results show that FCA outperforms existing methods across various test cases and generalizes well to different environments, vehicles, and target detectors.
Duan et al. proposed CAC \cite{CAC}, which conducts camouflage transformations in 3D space based on object pose changes. Their approach fixes top n proposals of the region proposal network and simultaneously attacks all classifications in fixed dense proposals to output errors. Experimental evaluations demonstrate CAC's superior performance in both virtual scenes and real-world settings compared to existing attack algorithms.

\textbf{Light-based physical attacks.} Light-based physical attacks utilize light beams as physical perturbations, projecting optimized beams onto target objects to conduct physical adversarial attacks on deep learning models. Duan et al. introduced AdvLB \cite{AdvLB}, employing a laser beam to execute physical adversarial attacks against deep learning models. They utilized a random search strategy to explore the physical parameters of the laser beam, achieving effective adversarial attacks in the physical world.
Hu et al. proposed AdvLS \cite{AdvLS}, which employs laser points as perturbations to perform covert attacks against Deep Neural Networks (DNNs). They utilized genetic algorithms to optimize the physical parameters of these laser points. Experimental results demonstrate AdvLS's efficacy in performing effective adversarial attacks against advanced DNNs.
Gnanasambandam et al. introduced OPAD \cite{OPAD}, an attack system comprising a low-cost projector, camera, and computer. This method incorporates the projection-camera model into adversarial attack optimization, developing novel attack paradigms capable of performing optical attacks on real 3D objects in the presence of background illumination.
Hu et al. proposed AdvCP \cite{AdvCP}, employing a projector to project a monochromatic beam onto the target object to execute physical attacks on the classifier, resulting in enhanced covert effects.

\subsection{Physical attacks in the infrared field}

Zhu et al. were pioneers in exploring physical attacks within the realm of infrared imaging, introducing the Bulb attack \cite{ref13}. This method utilized bulb heat sources to create white disturbance effects under infrared imaging, optimizing the position of each bulb through model gradients to deceive infrared imaging detectors. However, this approach required the pedestrian to carry the bulbs, lacking sufficient covertness and failing to execute attacks at varying distances.

Subsequently, Zhu et al. proposed QR attack \cite{ref14}, leveraging thermal insulation materials to optimize adversarial patterns in the digital environment, ultimately creating infrared adversarial clothing through manual simulation. Yet, the manual simulation process incurred high costs and resulted in conspicuous clothing patterns, easily detectable.

Wei et al. \cite{ref15} introduced AIP, which utilizes aerogel as a thermal insulation material. Adversarial patches are created via aggregation optimization techniques. In the physical realm, aerogel is precisely cut to match the shape of these patches and then affixed onto pedestrians for conducting physical attacks against infrared detectors. Notably, these methods constitute white-box attacks, necessitating access to internal information of the target model, rendering them impractical for real-world applications. 

To enhance camouflage effectiveness, Wei et al. devised HCB \cite{ref16}, utilizing hot and ice patches as physical disturbances to manifest white and black block effects under infrared imaging. The PSO algorithm optimized parameters for black-box attacks against infrared imaging detectors. However, HCB's design, based on a grid of nine squares, constrained query space, hindering efficient black-box query attacks.

Finally, Zhu et al. introduced AdvCloth \cite{AdvCloth}, elevating adversarial effects in both digital and physical realms by employing electrothermal patches as physical perturbations and the L\_dist loss function to constrain perturbations. Nonetheless, this method utilized white-box optimization, often inaccessible to attackers seeking internal information of the target model in practice.

\section{Methodology}

\subsection{Problem definition}

\textbf{Object detector.} In an infrared pedestrian dataset $D$, where $X$ denotes the clean image and $Y$ represents the correct label of the infrared pedestrian, while $f$ represents the object detector. For each input image X within the dataset D, the pretrained model $f: X \rightarrow Y$ of the object detector predicts a label $y$ that corresponds to the correct label $Y$. This predicted label $y$ comprises the location information ${y}_{pos}$, the target probability ${y}_{obj}$, and the target class ${y}_{cls}$:

\begin{equation}
    \label{Formula 1}
    y:=[{y}_{pos},{y}_{obj},{y}_{cls}]=f(X)
\end{equation}

\textbf{Particle Swarm Optimization (PSO):} PSO \cite{ref38} is an optimization technique inspired by the collective behavior of swarm creatures like birds or fish schools. In PSO, the search for an optimal solution in the solution space is simulated by mimicking the cooperation and information exchange observed in these swarms. Here's why PSO is advantageous for optimizing AdvIG:
(1) Simplicity and Ease of Implementation: The PSO algorithm is straightforward and easy to grasp, making it accessible even to beginners. Its simplicity translates into ease of implementation, requiring no complex mathematical expertise.
(2) Independence from Gradient Information: Unlike traditional optimization methods like gradient descent, PSO doesn't rely on gradient information from the objective function. This makes it particularly useful for problems where gradient information is unavailable or challenging to compute.
(3) Versatility: PSO demonstrates versatility in solving various optimization problems, including continuous, discrete, and multimodal optimization. It's applicable across different domains, encompassing function optimization, combinatorial optimization, parameter optimization, and more.
(4) Global Search Capability: Leveraging swarm intelligence, PSO constantly updates the positions and velocities of particles, enabling robust global search capabilities. It can explore the solution space efficiently, avoiding local optima and seeking better solutions effectively.

\subsection{Adversarial attack framework}

\begin{figure*}
\centering
\includegraphics[width=1\linewidth]{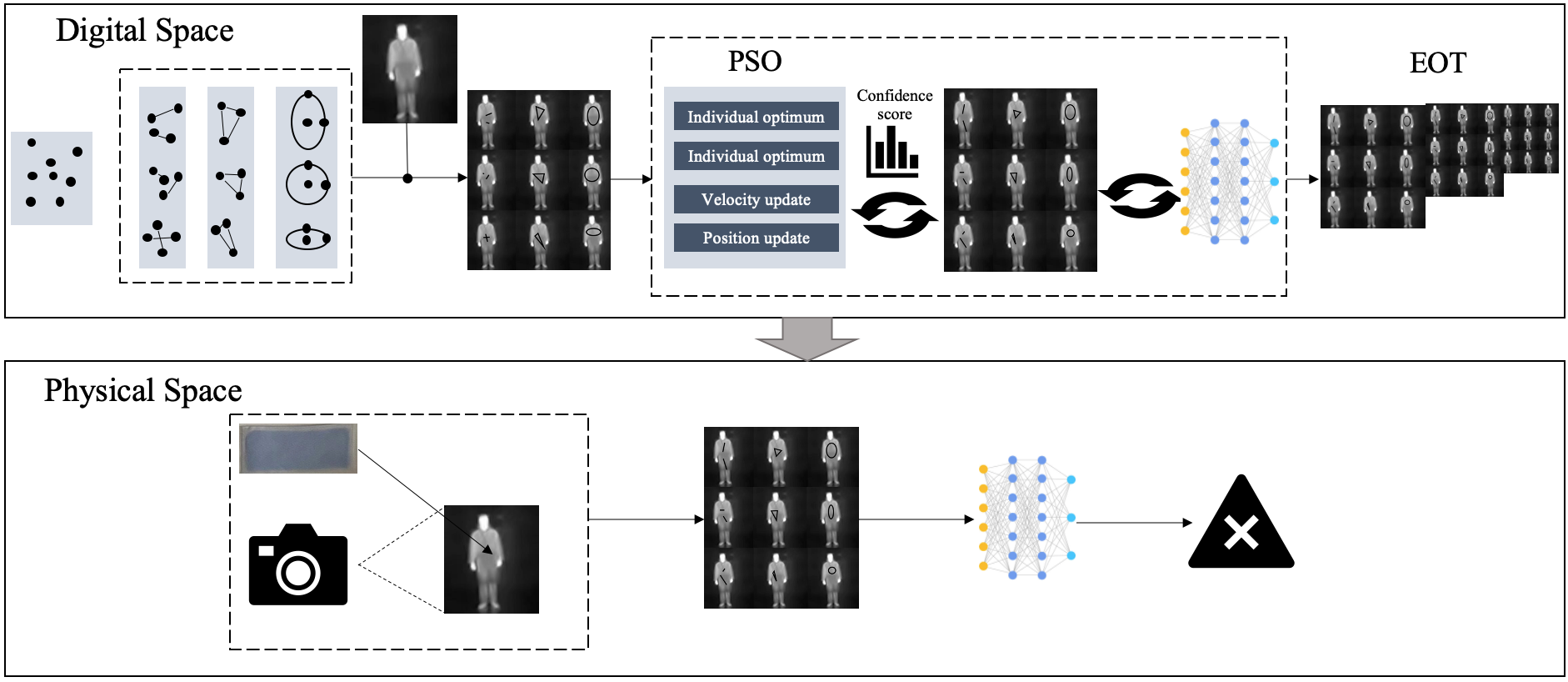} 
\caption{Adversarial attack framework of AdvIG.}
\label{figure2}
\end{figure*}

Figure \ref{figure2} illustrates the adversarial attack framework employed by the proposed AdvIG. It comprises four primary steps:
(1) Geometry Simulation and Modeling (Digital Environment): Initially, the geometry is simulated and modeled in the digital environment. Subsequently, it is integrated with clean samples to generate digital adversarial samples.
(2) Particle Swarm Optimization: PSO \cite{ref38} is utilized for adversarial optimization, iteratively refining the digital adversarial samples to obtain the most adversarial instances.
(3) Expectation Over Transformation (EOT) Framework: The EOT \cite{ref26} framework is applied to robustly enhance the generated digital adversarial examples, ensuring their effectiveness across various transformations.
(4) Physical Deployment (Physical World): In the physical world, the cold patch is affixed inside the pedestrian's clothing, and physical samples are captured using an infrared sensor.

\begin{figure*}
\centering
\includegraphics[width=1\linewidth]{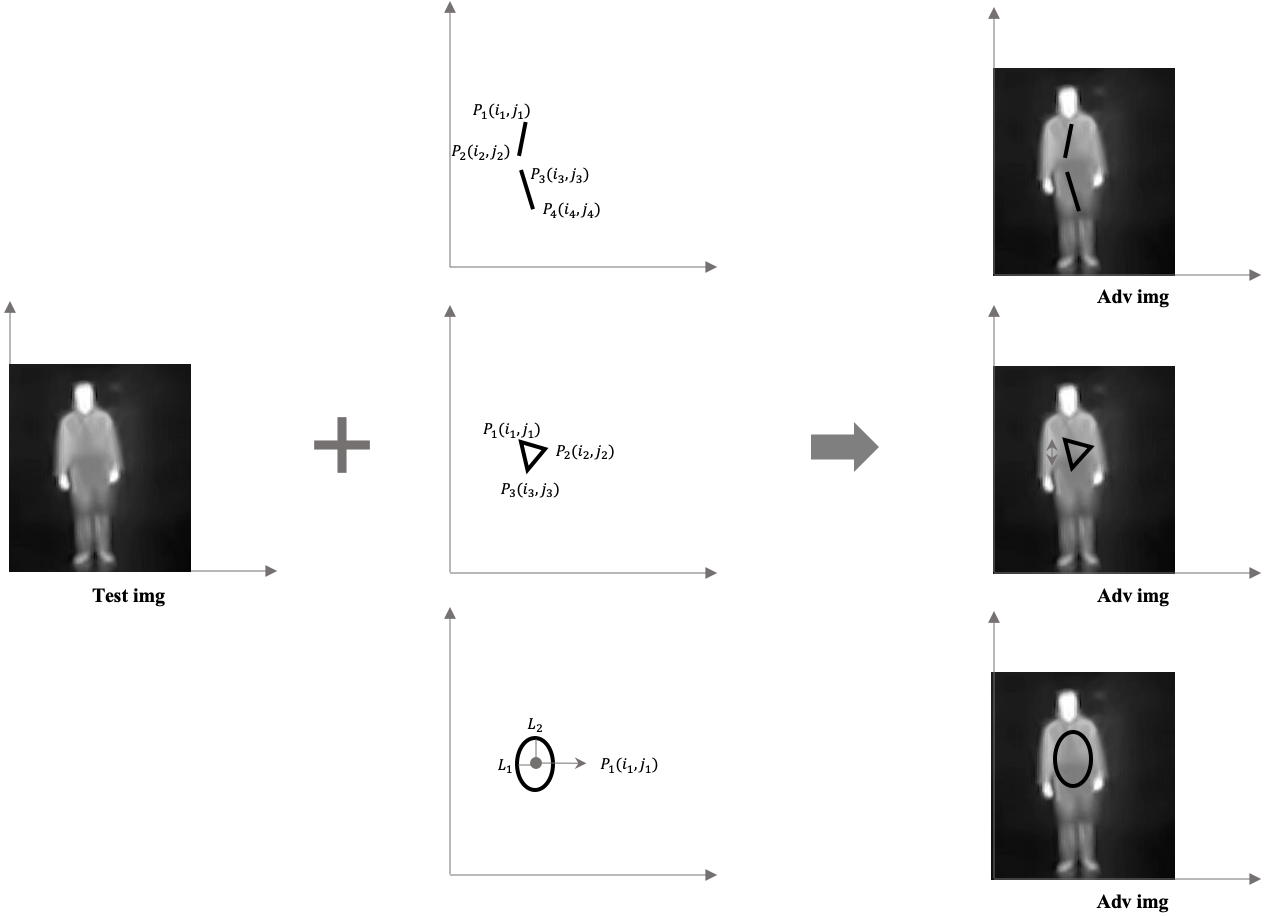} 
\caption{Schematic diagram of infrared geometry simulation modeling.}
\label{figure3}
\end{figure*}

\subsection{Infrared geometry modeling}

The simulation and modeling of infrared geometry are elaborated in Figure \ref{figure3}, showcasing the methods of line, triangle, and ellipse modeling. Each modeling technique is detailed below:

(1) Line Modeling:
To strike a balance between stealthiness and robustness, the line attack experiment employs two lines. Consequently, four point coordinates are employed to delineate the position of these lines, denoted as ${P}_{L} = {{P}_{1} ({i}_{1}, {j}_{1}), {P}_{2} ({i}_{2}, {j}_{2}), {P}_{3} ({i}_{3}, {j}_{3}), {P}_{4} ({i}_{4}, {j}_{4})}$. Here, ${i}_{k}$ and ${j}_{k}$ (where $k$ = 1, 2, 3, 4) denote the horizontal and vertical coordinates of the $k$-th vertex, respectively.

(2) Triangle Modeling:
The simulation of the triangle involves utilizing three vertices' position coordinates, expressed as ${P}_{T} = {{P}_{1} ({i}_{1}, {j}_{1}), {P}_{2} ({i}_{2}, {j}_{2}), {P}_{3} ({i}_{3}, {j}_{3})}$. Here, ${i}_{k}$ and ${j}_{k}$ (where $k$ = 1, 2, 3) represent the horizontal and vertical coordinates of the $k$-th vertex, respectively.

(3) Ellipse Modeling:
The ellipse modeling process incorporates three parameters, including the ellipse center coordinate ${P}_{E} = {{P}_{1} ({i}_{1}, {j}_{1})}$, the length of the horizontal axis (${L}_{1}$), and the length of the vertical axis (${L}_{2}$).

\subsection{Generate adversarial sample}

In this study, we introduce the mask $\mathcal{M}$ to confine the positioning of infrared geometric patterns within the pedestrian area, thereby preventing their extension beyond this designated region. Consequently, within the digital environment, the representation of digital adversarial samples is as follows:

\begin{equation}
    \label{Formula 2}
    {X}_{adv}^{L}=S(X, {P}_{L} \cap \mathcal{M})
\end{equation}

\begin{equation}
    \label{Formula 3}
    {X}_{adv}^{T}=S(X, {P}_{T} \cap \mathcal{M})
\end{equation}

\begin{equation}
    \label{Formula 4}
    {X}_{adv}^{E}=S(X, ({P}_{E},{L}_{1}, {L}_{2}) \cap \mathcal{M})
\end{equation}

Here, ${X}_{adv}^{L}$, ${X}_{adv}^{T}$, and ${X}_{adv}^{E}$ denote the adversarial samples generated using lines, triangle, and ellipse, respectively. $S$ represents a straightforward linear fusion function employed to amalgamate the generated infrared perturbations with the clean samples, thereby producing the final adversarial samples.

In order to mitigate experimental errors during the transition from digital to physical attacks, this study incorporates the Expectation Over Transformation (EOT) framework \cite{ref26}. EOT serves as an effective approach for navigating the shift from the digital domain to the physical domain in adversarial attacks. At its core, the framework involves transforming the original input data into samples with desired characteristics through an anticipated transformation function. This ensures that the model generates identical prediction results for both the original input and the transformed sample. The design of the desired transformation function typically considers the generation method of adversarial examples and the model's architecture, with the aim of preserving the naturalness and transferability of adversarial examples to the greatest extent possible.
The framework utilizes the distribution of a transformation $\mathcal{T}$ to model domain transfer, effectively addressing issues such as reduced robustness resulting from scaling. Specifically, for samples successfully executing adversarial attacks, they undergo multi-scale transformations. If the adversarial sample maintains its effectiveness under each scale transformation, it is deemed robust. Consequently, adversarial examples in the physical environment can be expressed as follows:

\begin{equation}
    \label{Formula 5}
    {X}_{adv}^{L}={\mathbb{E}}_{t \sim \mathcal{T}}t(S(X, {P}_{L} \cap \mathcal{M}))
\end{equation}

\begin{equation}
    \label{Formula 6}
    {X}_{adv}^{T}={\mathbb{E}}_{t \sim \mathcal{T}}t(S(X, {P}_{T} \cap \mathcal{M}))
\end{equation}

\begin{equation}
    \label{Formula 7}
    {X}_{adv}^{E}={\mathbb{E}}_{t \sim \mathcal{T}}t(S(X, ({P}_{E},{L}_{1}, {L}_{2}) \cap \mathcal{M}))
\end{equation}

\subsection{Infrared geometry adversarial attack}

This study focuses on black-box attack scenarios, employing adversarial attacks by querying the physical parameters of infrared geometries. Our objective is to identify the physical parameter $\theta$ ($\theta$ denotes either ${P}_{L}$, ${P}_{T}$, or $({P}_{E},{L}_{1},{L}_{2})$) of the most adversarial infrared geometry. This parameter is then utilized to simulate digital infrared perturbation, rendering the pedestrian undetectable by the infrared pedestrian detector in the disturbed image. We address the more realistic and challenging black-box attack scenario, where detailed information regarding the model's architecture and parameters is unavailable. Instead, only the input image and detection information of the model output (i.e., ${y}_{pos}$, ${y}_{obj}$, ${y}_{cls}$) are accessible. Consequently, we utilize the model output's probability ${y}_{obj}$ as the adversarial loss and formulate the optimization objective of the method as the minimization of ${y}_{obj}$:

\begin{equation}
    \label{Formula 8}
    \mathop{\arg\min}_{{P}_{L}}{\mathbb{E}}_{t \sim \mathcal{T}}({y}_{obj} \leftarrow t(f({X}_{adv}^{L})))
\end{equation}

\begin{equation}
    \label{Formula 9}
    \mathop{\arg\min}_{{P}_{T}}{\mathbb{E}}_{t \sim \mathcal{T}}({y}_{obj} \leftarrow t(f({X}_{adv}^{T}))
\end{equation}

\begin{equation}
    \label{Formula 10}
    \mathop{\arg\min}_{({P}_{E},{L}_{1},{L}_{2})}{\mathbb{E}}_{t \sim \mathcal{T}}({y}_{obj} \leftarrow t(f({X}_{adv}^{E}))
\end{equation}

In black-box attack scenarios, employing zeroth-order optimization \cite{ZOO} to estimate the gradient of the target model is a common approach. However, incorporating ZOO into the proposed method would encounter the issue of gradient explosion. This issue primarily arises due to the Boolean nature of determining whether the infrared pattern lies within the mask $\mathcal{M}$ region, rendering the use of ZOO impractical in this context. Consequently, we shift our focus to utilizing Particle Swarm Optimization (PSO) to optimize the physical parameters of infrared blocks. The process of optimizing AdvIG using PSO is outlined as follows:

(1) Initialization

In the PSO optimization process, initialization begins by randomly generating a population of candidate solutions. This population, denoted as $POP$, consists of individuals representing potential solutions. Each individual is associated with a velocity vector $V$:

\begin{equation}
    \label{Formula 11}
    POP=[{\theta}_{1},{\theta}_{2},...,{\theta}_{\alpha}]
\end{equation}

\begin{equation}
    \label{Formula 12}
    V=[{v}_{1},{v}_{2},...,{v}_{\alpha}]
\end{equation}

Where $\alpha$ denotes the population size, ${\theta}_{a}$ represents each candidate solution in the population $POP$, a ranges from 1 to $\alpha$. ${v}_{a}$ represents the direction in which the particle ${\theta}_{a}$ moves.

(2) Generate adversarial examples 

After random initialization, adversarial samples are generated for each individual ${\theta}_{a}$ in the population $POP$ using equations \ref{Formula 2}, \ref{Formula 3}, and \ref{Formula 4}:

\begin{equation}
    \label{Formula 13}
    {X}_{a}^{b}=S(X,{\theta}_{a} \cap \mathcal{M})
\end{equation}

Where $b$ represents the current iteration number of the population. ${X}_{a}^{b}$ denotes the adversarial sample generated by the $a$-th individual in the population of generation $b$.

3) Obtain the individual optimal solution and the global optimal solution

By obtaining the individual optimal solution and the global optimal solution, they are used to guide the group evolution direction:

\begin{equation}
    \label{Formula 14}
    {\theta}_{a,best}^{b}=\mathop{\arg\min}_{{\theta}_{a,best}^{u}}{y}_{obj} \leftarrow f({X}_{a}^{u})  \quad u \in [1, b]
\end{equation}

\begin{equation}
    \label{Formula 15}
    {\theta}_{best}^{b}=\mathop{\arg\min}_{{\theta}_{a,best}^{u}}{y}_{obj} \leftarrow f({X}_{a}^{u}) \quad a \in [1, \alpha] , u \in [1, b]
\end{equation}

Where ${\theta}_{a,best}^{b}$, ${\theta}_{best}^{b}$ represent the optimal solution of the $a$-th individual in the $b$-generation population and the global optimal solution of the population, respectively.

\begin{algorithm}[b]
	\renewcommand{\algorithmicrequire}{\textbf{Input:}}
	\renewcommand{\algorithmicensure}{\textbf{Output:}}
	\caption{Pseudocode of AdvIG}
	\label{algorithm1}
	\begin{algorithmic}[1]
	
		\REQUIRE Clean sample $X$, Detector $f$, Population size $\alpha$, Iterations $I$, Hyperparameters of PSO: $\omega$,${c}_{1}$,${r}_{1}$,${c}_{2}$, ${r}_{2}$;
		\ENSURE Physical parameters ${\theta}^{\star}$;

		\STATE \textbf{Initialization} Randomly set $POP$, $V$;
        \FOR{$b$ $\leftarrow$ 0 to $I$}
            \FOR{each ${\theta}_{a}^{b}$ in ${POP}^{b}$}
                \STATE ${X}_{a}^{b}=S(X,{\theta}_{a}^{b} \cap \mathcal{M})$;
                \STATE ${\theta}_{a,best}^{b}=\mathop{\arg\min}_{{\theta}_{a,best}^{u}}{y}_{obj} \leftarrow f({X}_{a}^{u})  \quad u \in [1, b]$;
                \STATE ${\theta}_{best}^{b}=\mathop{\arg\min}_{{\theta}_{a,best}^{u}}{y}_{obj} \leftarrow f({X}_{a}^{u}) \quad a \in [1, \alpha] , u \in [1, b]$;
                \STATE ${\theta}^{\star}={\theta}_{best}^{b}$;      
            \ENDFOR
            
            \STATE ${v}_{a}^{b+1}=\omega{v}_{a}^{b}+{c}_{1}{r}_{1}({\theta}_{a,best}^{b}-{\theta}_{a}^{b}) + {c}_{2}{r}_{2}({\theta}_{best}^{b}-{\theta}_{a}^{b})$;
            \STATE ${\theta}_{a}^{b+1}={\theta}_{a}^{b}+{v}_{a}^{b+1}$;
        \ENDFOR
        
        \STATE \textbf{Output:} ${\theta}^{\star}$;

	\end{algorithmic}  
\end{algorithm}

(4) Update the velocity and position information of the individual 

After obtaining the individual optimal solution and the global optimal solution, the information is used to guide the population towards the direction that is easier to obtain the optimal solution. The following formula is used to update the individual velocity direction and position of the population:

\begin{equation}
    \label{Formula 16}
    {v}_{a}^{b+1}=\omega{v}_{a}^{b}+{c}_{1}{r}_{1}({\theta}_{a,best}^{b}-{\theta}_{a}^{b}) + {c}_{2}{r}_{2}({\theta}_{best}^{b}-{\theta}_{a}^{b})
\end{equation}

\begin{equation}
    \label{Formula 17}
    {\theta}_{a}^{b+1}={\theta}_{a}^{b}+{v}_{a}^{b+1}
\end{equation}

Here, $\omega$, ${c}_{1}$, ${r}_{1}$, ${c}_{2}$, ${r}_{2}$, are the hyperparameters of PSO, and $\omega$ denotes the inertia factor. ${c}_{1}$ and ${c}_{2}$ denote the learning factors of the particles. ${r}_{1}$ and ${r}_{2}$ are random numbers generated from a uniform distribution in the range [0,1].

Algorithm \ref{algorithm1} presents the pseudocode for AdvIG, detailing its optimization process. This algorithm requires a clean sample $X$, a target detector $f$, parameters such as population size $\alpha$ and iteration number $I$, and PSO hyperparameters $\omega$, ${c}_{1}$, ${r}_{1}$, ${c}_{2}$, ${r}_{2}$ as inputs. It outlines the steps for optimizing the physical parameters of the most adversarial infrared perturbation. This optimal solution is then utilized to execute subsequent physical attack experiments.

\section{Experiments}
\label{sec4}

\subsection{Experimental setting}

\textbf{Dataset:} In line with the approach adopted in Bulb attack, we utilize the FLIR\_ADAS v1\_3 dataset provided by FLIR company for our experiments. The FLIR dataset is extensively utilized in the research and development of infrared image processing and computer vision algorithms. It primarily serves the purpose of object detection and tracking tasks, particularly focusing on human target detection and tracking in infrared images. This dataset comprises a diverse array of infrared images captured across various scenes and environments, including indoor, outdoor, daytime, and nighttime scenarios. Each image in the FLIR dataset comes with corresponding annotations specifying the location and category of objects, such as pedestrians, vehicles, and others.
For our proposed AdvIG method, we specifically target the infrared pedestrian detector. To this end, we filter the original dataset to retain only pedestrian instances with a height greater than 120 pixels. Subsequently, we obtain a subset consisting of 1011 images. From this filtered dataset, 710 images are allocated for training purposes, while the remaining 301 images are reserved for testing.

\textbf{Object Detector:}  Similar to Bulb attack, we employ the Yolo v3 \cite{ref4} detector as our target detector for the infrared pedestrian detector. To expedite convergence, we fine-tune the Yolo v3 detector on the dataset using pre-trained weights. Upon training, the final infrared pedestrian detector achieved an average accuracy of 90.1\% on the training set and 90.7\% on the test set.

\begin{figure}
\centering
\includegraphics[width=0.7\columnwidth]{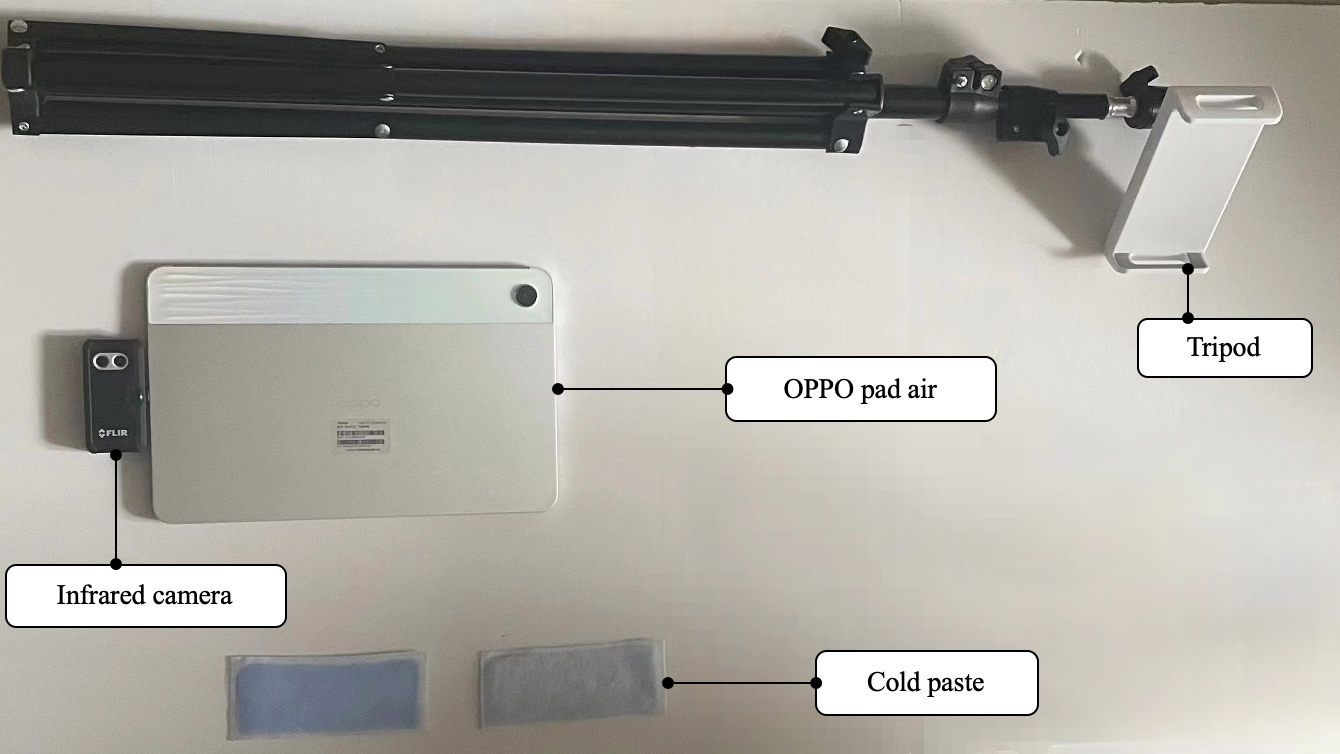} 
\caption{Experimental devices. The experimental devices for AdvIG include a FLIR one infrared camera, an OPPO tablet, a tripod, and cold pastes.}
\vspace{-0.3cm}
\label{figure4}
\end{figure}

\textbf{Experimental devices:}  The experimental devices utilized in this study is illustrated in Figure \ref{figure4} and comprises essential components such as a tripod, a FLIR One infrared camera, an OPPO tablet, and several cold patches. The specifications of the infrared camera are as follows: infrared resolution of 160×120, visible resolution of 1440×1080, and an infrared NETD of less than 70mk. The cold patches used in the experiments are capable of maintaining a temperature of 24 degrees Celsius for over four hours.

\textbf{Evaluation Criteria:} The primary objective of this study is to execute the vanish attack, wherein the object detector fails to detect pedestrians post perturbation. Consequently, we employ the attack success rate (ASR) as the metric to assess the adversarial efficacy of the proposed AdvIG. A higher ASR signifies a more potent AdvIG, while a lower rate indicates its ineffectiveness. The ASR is calculated as follows:

\begin{equation}
\label{Formula 18}
\begin{split}
    &{\rm ASR}(X) = 1-\frac{1}{N}\sum_{n=1}^{N}F({y}_{obj}^{n})\\
    &F({y}_{obj}^{n})=
        \begin{cases}
        0 & {y}_{obj}^{n} < 0.5 \\
        1 & otherwise
        \end{cases}
\end{split}
\end{equation}

Here, $N$ denotes the total count of positive tags in the clean dataset that can be detected by the infrared target detector. Throughout all attack experiments conducted in this study, the threshold is uniformly set to 0.5. This signifies that the vanishing attack is deemed successful when the confidence level of the detected target falls below this threshold.

\textbf{Competing Methods:} Our proposed AdvIG operates within the framework of black-box attacks. To ensure a fair comparison, we select the black-box attack method, HCB \cite{ref16}, as the baseline for experimental evaluation. To maintain a fair comparison, in the digital/physical attack experiments, the number of perturbations generated by HCB is no fewer than that of our method. Additionally, the same object detection model is employed to further ensure the fairness of the comparison.

\textbf{Additional Details:} The hyperparameters of PSO are configured as follows: $\omega$ = 0.9, ${c}_{1}$ = 1.6, ${r}_{1}$ = 0.5, ${c}_{2}$ = 1.4, ${r}_{2}$ = 0.5. All attack experiments are conducted on a single NVIDIA GeForce RTX 4090 GPU.

\begin{figure}[b]
\centering
\includegraphics[width=1\columnwidth]{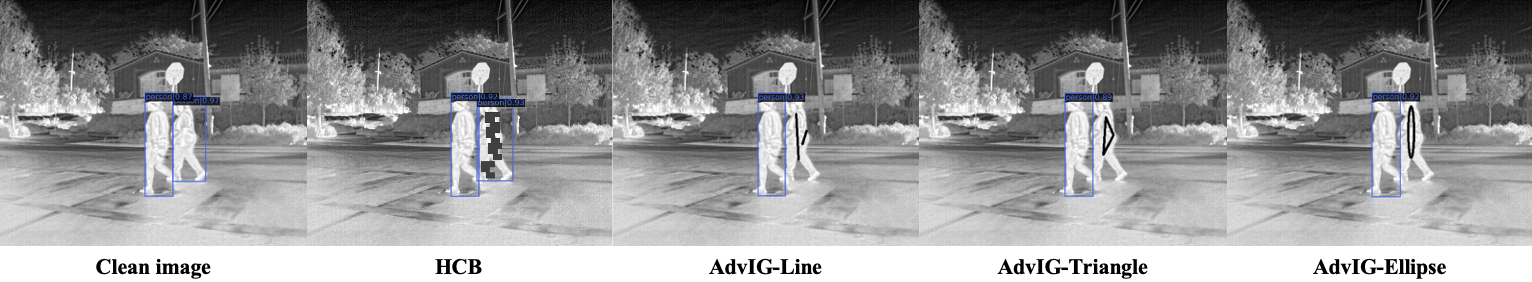} 
\caption{Digital samples of AdvIG.}
\vspace{-0.3cm}
\label{figure5}
\end{figure}

\subsection{Evaluation of effectiveness}

\textbf{Digital attacks.} In the digital attack phase, we employ two lines, a triangle, and an ellipse to execute adversarial attacks on the infrared pedestrian detector, with the experimental outcomes summarized in Table \ref{Table2}. The results demonstrate that our proposed AdvIG consistently achieves an attack success rate of no less than 85\% against the infrared pedestrian detector in the digital environment. Notably, utilizing the ellipse to execute the infrared adversarial attack yields the maximum adversarial effect, achieving a remarkable 100\% attack success rate. Furthermore, the average number of queries required for utilizing lines, triangles, and ellipses is 71.1, 113.1, and 2.6, respectively, underscoring the efficiency of AdvIG. Figure \ref{figure5} showcases the digital adversarial samples generated by our method, illustrating that AdvIG produces adversarial samples with minimal perturbation compared to HCB, while still achieving effective adversarial attacks.

\begin{table} 
	\centering
 
    \setlength{\belowcaptionskip}{5pt}
    \caption{Experimental results of digital attacks.}
    \label{Table2}
	\begin{tabular}{cccc}

    \hline
    ~&Line&Triangle&Ellipse\\
    \hline
    ASR (\%)&91.3&86.8&100\\
    \hline
    Query&71.7&133.1&2.6\\
    \hline

\end{tabular}
\end{table}

\begin{table} 
	\centering
 
    \setlength{\belowcaptionskip}{5pt}
    \caption{Experimental results of physical attacks (\%).}
    \label{Table3}
	\begin{tabular}{ccccccc}

    \hline
    ~&6m&5m&4m&3m&2m&Avg\\
    \hline
    Line&66.7&97.7&59.7&44.3&37.2&61.1\\
    \hline
    Triangle&88.9&73.5&63.9&33.8&45.8&61.2\\
    \hline
    Ellipse&91.2&91.7&98.3&100.0&100.0&96.2\\
    \hline
    Line+Triangle&96.7&53.2&67.4&40.3&23.8&56.3\\
    \hline
    Line+Ellipse&85.3&77.8&59.7&25.5&50.0&59.7\\
    \hline
    Triangle+Ellipse&66.7&57.4&40.5&51.3&17.1&46.6\\
    \hline
    Line+Triangle+Ellipse&83.3&85.6&66.7&18.9&25.0&55.9\\
    \hline

\end{tabular}
\end{table}

\begin{figure}
\centering
\includegraphics[width=1\columnwidth]{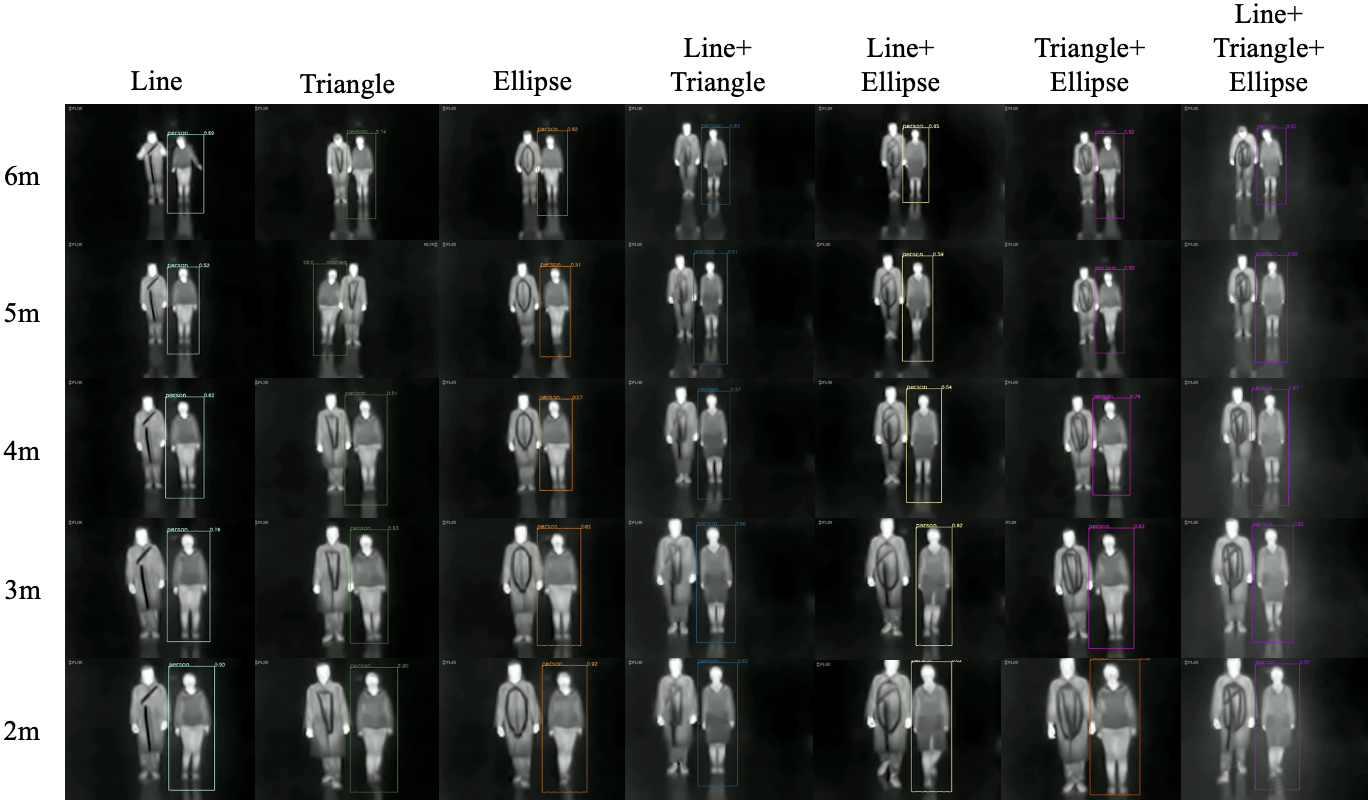} 
\caption{Physical samples of AdvIG.}
\vspace{-0.3cm}
\label{figure6}
\end{figure}

\textbf{Physical attacks.} In the physical attack phase, we aim to comprehensively assess the adversarial effect of AdvIG across varying distances, ranging from 2 meters to 6 meters with 1-meter intervals. Employing lines, triangles, and ellipses separately, as well as their combined attacks, we present the experimental results in Table \ref{Table3}, from which nine key observations can be drawn:
(1) Overall, the average attack success rates of 61.1\%, 61.2\%, 96.2\%, 56.3\%, 59.7\%, 46.6\%, and 55.9\% are achieved by using lines, triangles, ellipses, line + triangle, line + triangle + ellipse, and line + triangle + ellipse, respectively, with ellipses demonstrating the strongest attack effect, while the combination of triangle + ellipse exhibits the weakest adversarial effect.
(2) The use of infrared geometry for physical attacks reveals that the adversarial effect of combined graphics attacks is not necessarily superior to that of individual graphics. Notably, the attack success rate of the ellipse attack surpasses that of the combination of ellipse with other shapes.
(3) Regarding line attacks, the optimal adversarial effect is observed at a distance of 5 meters, with the success rate decreasing as the distance either increases or decreases.
(4) For triangle attacks, the highest success rate is achieved at a distance of 6 meters, with success rates exhibiting a downward trend as the distance decreases.
(5) Elliptical attacks show an increasing success rate trend with decreasing distance, reaching a maximum of 100\% at 3 meters.
(6) The line + triangle combination attack achieves maximum success at 6 meters, with success rates declining as the distance decreases.
(7) In the line + ellipse combination attack, the highest success rate is observed at 6 meters, declining as the distance decreases, reaching a minimum at 3 meters, before increasing again.
(8) With the triangle and ellipse combination attack, the highest success rate is observed at 6 meters, with subsequent success rates exhibiting fluctuation.
(9) The combination of line, triangle, and ellipse attacks demonstrates an overall declining success rate trend, with the highest success rate at 6 meters and the lowest at 3 meters.
Figure \ref{figure6} illustrates the physical adversarial samples generated by AdvIG, showcasing the effectiveness of using different adversarial infrared geometries, as well as combined graphics, to achieve effective adversarial attacks on the infrared pedestrian target detector across varying distances.

\subsection{Evaluation of stealthiness}

As previously discussed, the prevailing infrared attack methods \cite{ref13,ref14,ref15} typically apply perturbations externally, making them easily detectable by human observers. In contrast, AdvIG strategically positions the perturbation within the clothing, rendering it challenging for human observers to discern its presence without the aid of infrared sensors.

Comparing the stealthiness of AdvIG with HCB \cite{ref16}, in the digital environment, as depicted in Figure \ref{figure5}, the perturbations generated by AdvIG exhibit subtler characteristics compared to HCB, resulting in a more covert effect. In the physical environment, Figure \ref{figure1} illustrates that AdvIG manifests fewer perturbations, further enhancing its stealthiness relative to HCB.

Consequently, the proposed AdvIG method surpasses the baseline approach in terms of stealthiness, primarily owing to its adept utilization of internal perturbation placement, contributing to reduced visibility and enhanced covert operation.

\subsection{Evaluation of robustness}

We subject the proposed AdvIG to a series of black-box attacks against advanced detectors to assess its robustness, including DETR \cite{ref41}, Mask Rcnn \cite{ref42}, Faster Rcnn \cite{ref43}, Libra Rcnn \cite{ref44}, and RetinaNet \cite{ref45}. These pre-trained models are fine-tuned on the filtered infrared dataset, culminating in average accuracies of 91.2\%, 89.5\%, 90.8\%, 88.0\%, and 93.0\%, respectively. The experimental findings are summarized in Table \ref{Table4}, yielding the following conclusions:
(1) Overall, employing line, triangle, and ellipse for digital attacks yields average attack success rates of 82.4\%, 81.7\%, and 92.5\% against the advanced target detectors, respectively. 
(2) The deployment of line, triangle, and ellipse in digital attacks requires an average number of queries less than 150, underscoring the efficiency of AdvIG.
(3) Among the tested detectors, DETR demonstrates a more robust adversarial effect against AdvIG across all attack experiments, suggesting that Transformer-based detector exhibit better resilience to adversarial attacks.

\begin{table} 
	\centering
 
    \setlength{\belowcaptionskip}{5pt}
    \caption{Deploy AdvIG to attack various infrared pedestrian detectors.}
    \label{Table4}
	\begin{tabular}{ccccccccc}

    \hline
    ~&~&Yolo v3&DETR&Mask&Faster&Libra&Retina&avg\\
    \hline
    \multirow{2}*{Line}&ASR&93.1&42.4&90.2&96.2&98.4&74.3&82.4\\
    \cmidrule(r){2-9}
    ~&Query&71.7&350.8&98.8&46.2&49.5&157.8&129.1\\
    \hline
    \multirow{2}*{Triangle}&ASR&86.8&41.4&92.2&96.2&96.8&77.1&81.7\\
    \cmidrule(r){2-9}
    ~&Query&113.1&356.2&104.9&38.5&56.0&150.7&136.6\\
    \hline
    \multirow{2}*{Ellipse}&ASR&100.0&58.6&98.1&99.1&99.2&100.0&92.5\\
    \cmidrule(r){2-9}
    ~&Query&2.6&218.6&13.3&6.7&6.5&3.2&41.8\\
    \hline

\end{tabular}
\end{table}

We reproduce HCB \cite{ref16} and present its experimental results in Table \ref{Table5} alongside those of our proposed approach. HCB utilizes 14 cold patches in the physical experiments, whereas our method employs fewer cold patches in the physical experiments compared to HCB. However, despite using fewer cold patches, our method demonstrates the ability to execute more robust physical attacks.

\begin{table} 
	\centering
 
    \setlength{\belowcaptionskip}{5pt}
    \caption{Comparison of experimental results between the proposed AdvIG and baseline method.}
    \label{Table5}
	\begin{tabular}{ccccc}

    \hline
    \multirow{2}*{Method} & \multicolumn{2}{c}{Digital} & \multicolumn{2}{c}{Physical }\\
    \cmidrule(r){2-3}
    \cmidrule(r){4-5}
    ~&ASR (\%)&Query&ASR (\%)&Infrared patches\\
    \hline
    HCB&37.3&694.0&51.6&14\\
    \hline
    AdvIG-Line&93.1&71.7&61.1&\textbf{8}\\
    \hline
    AdvIG-Triangle&86.8&113.1&61.2&10\\
    \hline
    AdvIG-Ellipse&\textbf{100.0}&\textbf{2.6}&\textbf{96.2}&12\\
    \hline

\end{tabular}
\end{table}

In summary, the experimental findings presented in Table \ref{Table4} demonstrate the efficacy of the proposed AdvIG in executing effective adversarial attacks against various advanced object detectors. Moreover, the results in Table \ref{Table5}  illustrate that AdvIG outperforms the baseline method, underscoring the robustness and superiority of the proposed approach.

\section{Discussion}

\subsection{Deploy AdvIG to attack infrared vehicle detectors}

In addition to assessing the adversarial impact of AdvIG on infrared pedestrian detectors, we extend its application to infrared vehicle detectors. We train infrared vehicle detectors, including Yolo v3 \cite{ref4}, DETR \cite{ref41}, Mask Rcnn \cite{ref42}, Faster Rcnn \cite{ref43}, Libra Rcnn \cite{ref44}, and RetinaNet \cite{ref45}, on the FLIR dataset, achieving average accuracies of of 92.1\%, 94.6\%, 94.2\%, 94.4\%, 95.6\% and 95.5\%. Subsequently, AdvIG is deployed to attack these finely trained infrared vehicle detectors, with experimental outcomes summarized in Table \ref{Table6}. The following conclusions are drawn:
(1) Overall, employing line, triangle, and ellipse attacks against vehicle detectors resulte in effective adversarial effects, with achieved ASRs of 77.4\%, 82.2\%, and 50.4\%, respectively. Moreover, the average number of queries remained below 300, validating the robustness and efficiency of AdvIG in attacking vehicle detectors.
(2) Unlike the scenario with infrared pedestrian detectors, the performance of ellipse attacks is inferior to that of line and triangle attacks when attacking vehicle detectors.
(3) Similar to the case of attacking infrared pedestrian detectors, DETR exhibited more robust performance among the various detectors.

\begin{figure}
\centering
\includegraphics[width=1\columnwidth]{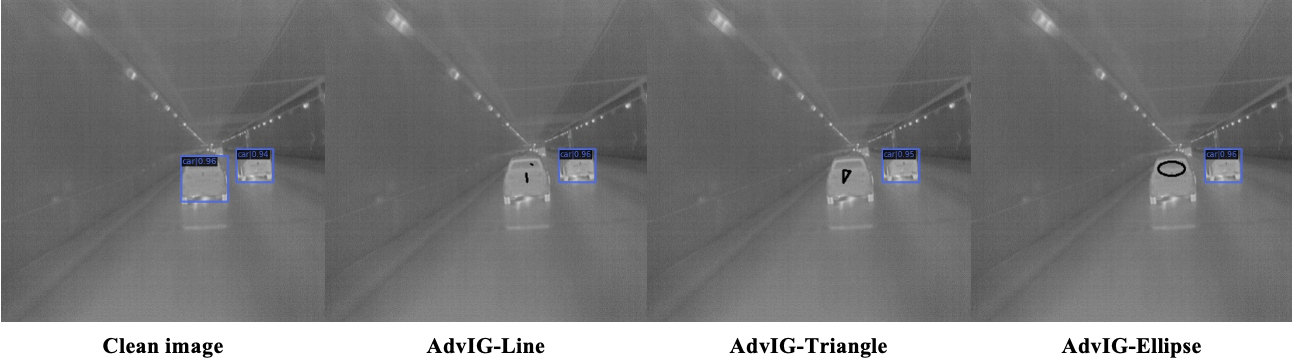} 
\caption{Demonstration of adversarial samples generated by deploying AdvIG to attack infrared vehicle detector.}
\vspace{-0.3cm}
\label{figure7}
\end{figure}

Furthermore, Figure \ref{figure7} illustrates the generated adversarial examples when attacking the Infrared Vehicle Object Detector (Yolo v3), demonstrating the successful execution of attacks on the infrared vehicle object detector using various infrared geometries.

\begin{table} 
	\centering
 
    \setlength{\belowcaptionskip}{5pt}
    \caption{Deploy AdvIG to attack various infrared vihecle detectors.}
    \label{Table6}
	\begin{tabular}{ccccccccc}

    \hline
    ~&~&Yolo v3&DETR&Mask&Faster&Libra&Retina&avg\\
    \hline
    \multirow{2}*{Line}&ASR&99.0&42.9&66.7&90.5&82.8&82.4&77.4\\
    \cmidrule(r){2-9}
    ~&Query&19.7&321.2&222.9&89.7&145.1&112.0&151.8\\
    \hline
    \multirow{2}*{Triangle}&ASR&100.0&47.1&71.8&93.8&92.2&88.5&82.2\\
    \cmidrule(r){2-9}
    ~&Query&10.0&306.9&194.6&68.9&87.2&84.3&125.3\\
    \hline
    \multirow{2}*{Ellipse}&ASR&97.1&7.1&25.6&52.9&38.5&81.3&50.4\\
    \cmidrule(r){2-9}
    ~&Query&18.8&471.8&387.3&243.5&316.3&100.0&256.3\\
    \hline

\end{tabular}
\end{table}

\subsection{Ablation study}

(1) Ablation study on the number of lines

In this experiment, we vary the number of straight lines from 1 to 7, with increments of 1, to investigate the influence of the number of lines on the adversarial attack effectiveness of AdvIG. The experimental results are summarized in Table \ref{Table7}. It is observed that the overall attack success rate demonstrates an increasing trend with the augmentation of the number of lines. However, when the number of lines is 2, the attack success rate reaches a threshold, indicating that beyond this point, the growth in attack success rate becomes marginal.

\begin{table} 
	\centering
 
    \setlength{\belowcaptionskip}{5pt}
    \caption{Ablation study on the number of lines.}
    \label{Table7}
	\begin{tabular}{cccccccc}
    
    \hline
    ~&1&2&3&4&5&6&7\\
    \hline
    ASR&81.6&93.1&92.0&96.0&96.6&96.0&96.6\\
    \hline
    Query&131.7&71.7&64.1&50.1&46.6&49.6&43.6\\

    \hline

\end{tabular}
\end{table}

(2) Polygon edge number ablation experiment

In this experiment, we vary the polygon's edge numbers from triangular to nonagon to investigate the influence of different polygon shapes on the adversarial effect of AdvIG. The experimental results are presented in Table \ref{Table8}, revealing that the attack success rate generally increases with the augmentation of the number of polygon edges. Interestingly, when the number of edges reaches 4, the attack success rate reaches a threshold. Additionally, it is noteworthy that robust adversarial attack effects can be achieved even with triangles.

\begin{table} 
	\centering
 
    \setlength{\belowcaptionskip}{5pt}
    \caption{Ablation study of polygon edge number.}
    \label{Table8}
	\begin{tabular}{cccccccc}
    
    \hline
    ~&3&4&5&6&7&8&9\\
    \hline
    ASR&86.8&91.4&90.8&93.1&94.8&93.7&93.7\\
    \hline
    Query&113.1&91.2&88.6&80.3&75.6&77.8&74.1\\

    \hline

\end{tabular}
\end{table}

(3) Ablation study of color

In this study, we examine the influence of different colors of infrared perturbations on the attack effectiveness by selecting black as the color for adversarial perturbations in digital and physical attack experiments. We also explore various color values, including (0,0,0), (51, 51), (102,102,102), (153, 153), (204,204,204), and (255,255,255). The experimental results are presented in Table \ref{Table9}, leading to the following conclusions: (1) Across all color variations, AdvIG consistently achieves an attack success rate of no less than 40\%, indicating its robust adversarial effect under different color conditions. (2) Notably, when the perturbation color is black (0,0,0), AdvIG exhibit the highest attack success rates, with line, triangle, and ellipse achieving success rates of 78.3\%, 86.8\%, and 100\%, respectively. (3) Conversely, the lowest attack success rates are observed when the color is (153,153,153), with attack success rates of 46.7\%, 51.1\%, and 80.0\%, respectively. This can be attributed to the color similarity between the infrared perturbation and the pedestrian in the dataset, resulting in a weaker disturbance effect.

\begin{table} 
	\centering
 
    \setlength{\belowcaptionskip}{5pt}
    \caption{Ablation study of color.}
    \label{Table9}
    \resizebox{\columnwidth}{!}{
	\begin{tabular}{cccccccc}

    \hline
    ~&~&(0,0,0)&(51,51,51)&(102,102,102)&(153,153,153)&(204,204,204)&(255,255,255)\\
    \hline
    \multirow{2}*{Line}&ASR&93.1&77.2&57.8&46.7&59.4&75.0\\
    \cmidrule(r){2-8}
    ~&Query&71.7&174.5&273.4&313.8&262.7&179.8\\
    \hline
    \multirow{2}*{Triangle}&ASR&86.8&71.7&57.8&51.1&60.6&78.3\\
    \cmidrule(r){2-8}
    ~&Query&113.1&200.1&282.7&310.5&261.8&180.9\\
    \hline
    \multirow{2}*{Ellipse}&ASR&100.0&98.9&88.9&80.0&87.8&95.6\\
    \cmidrule(r){2-8}
    ~&Query&2.6&9.0&60.9&106.6&65.9&26.3\\
    \hline

\end{tabular}
}
\end{table}

\subsection{Transferability of AdvIG}

(1) Digital transfer attack

In this experiment, we utilize the samples that effectively attacked Yolo v3 in the digital environment as the dataset, conducting transfer attacks on DETR, Mask Rcnn, Faster Rcnn, Libra Rcnn, and RetinaNet, respectively. The results are presented in Table \ref{Table10}, indicating a relatively weak digital transfer attack effect of AdvIG. In the most favorable scenario, AdvIG achieves a success rate of 56.1\% for digital transfer attacks.

\begin{table} 
	\centering
 
    \setlength{\belowcaptionskip}{5pt}
    \caption{Experimental results of digital transfer attacks.}
    \label{Table10}
	\begin{tabular}{cccccc}

    \hline
    ~&DETR&Mask&Faster&Libra&Retina\\
    \hline
    Line&5.4&13.7&20.8&21.4&31.5\\
    \hline
    Triangle&4.5&10.8&17.2&20.4&29.9\\
    \hline
    Ellipse&6.1&28.9&37.2&36.1&56.1\\
    \hline
    
\end{tabular}
\end{table}

(2) Physical transfer attack

In the physical transfer attack experiment, we utilize the physical samples that successfully attacked Yolo v3 at various distances as the attack dataset, and conduct transfer attacks on DETR, Mask Rcnn, Faster Rcnn, Libra Rcnn, and RetinaNet, respectively. The experimental results are presented in Tables \ref{Table11}, \ref{Table12}, \ref{Table13}, \ref{Table14}, \ref{Table15}, \ref{Table16}, \ref{Table17}. The following conclusions can be drawn:
1) Line physical transfer attack: When the distance is close, the deployment of line physical transfer attack demonstrates a certain adversarial effect, achieving a maximum attack success rate of 76.5\%. However, in some cases, the attack success rate is 0.
2) Triangle physical transfer attack: In more than half of the cases, the success rate is 0, with the highest success rate reaching 61.5\%.
3) Ellipse physical transfer attack: Similar to triangle attack, some cases result in a success rate of 0, while in the best case, the success rate reaches 79.7\%.
4) Line + triangle physical transfer attack: In the majority of cases, the attack success rate is 0, but in the best case, it reaches 83.3\%.
5) Line + ellipse physical transfer attack: While some cases have a success rate of 0, in most instances, the migration attack is effectively executed, with the highest success rate reaching 78.4\%.
6) Triangle + ellipse physical transfer attack: AdvIG demonstrates effective transfer attacks in nearly all cases, achieving a maximum success rate of 90.0\%.
7) Line + triangle + ellipse physical transfer attack: In rare cases, the success rate is 0, yet in the best case, the success rate reaches 100\%.

\begin{table} 
	\centering
 
    \setlength{\belowcaptionskip}{5pt}
    \caption{Experimental results of physical transfer attacks (Line).}
    \label{Table11}
	\begin{tabular}{cccccc}

    \hline
    ~&DETR&Mask&Faster&Libra&Retina\\
    \hline
    2m&23.5&47.1&56.9&76.5&2.0\\
    \hline
    3m&4.7&32.6&39.5&23.3&2.3\\
    \hline
    4m&0.0&8.1&0.0&32.4&40.5\\
    \hline
    5m&0.0&2.4&2.4&45.2&7.1\\
    \hline
    6m&0.0&0.0&0.0&0.0&12.5\\
    \hline
    
\end{tabular}
\end{table}

\begin{table} 
	\centering
 
    \setlength{\belowcaptionskip}{5pt}
    \caption{Experimental results of physical transfer attacks (Triangle).}
    \label{Table12}
	\begin{tabular}{cccccc}

    \hline
    ~&DETR&Mask&Faster&Libra&Retina\\
    \hline
    2m&28.9&39.5&39.5&57.9&2.6\\
    \hline
    3m&0.0&61.5&30.8&26.9&0.0\\
    \hline
    4m&0.0&13.0&0.0&17.4&0.0\\
    \hline
    5m&0.0&4.0&0.0&4.0&4.0\\
    \hline
    6m&0.0&0.0&0.0&0.0&1.6\\
    \hline
    
\end{tabular}
\end{table}

\begin{table} 
	\centering
 
    \setlength{\belowcaptionskip}{5pt}
    \caption{Experimental results of physical transfer attacks (Ellipse).}
    \label{Table13}
	\begin{tabular}{cccccc}

    \hline
    ~&DETR&Mask&Faster&Libra&Retina\\
    \hline
    2m&37.9&31.0&27.6&10.3&19.0\\
    \hline
    3m&14.9&59.7&25.4&38.8&0.0\\
    \hline
    4m&6.8&45.8&23.7&79.7&0.0\\
    \hline
    5m&0.0&27.3&1.8&38.2&16.4\\
    \hline
    6m&0.0&25.0&0.0&19.2&61.5\\
    \hline
    
\end{tabular}
\end{table}

\begin{table} 
	\centering
 
    \setlength{\belowcaptionskip}{5pt}
    \caption{Experimental results of physical transfer attacks (Line+Triangle).}
    \label{Table14}
	\begin{tabular}{cccccc}

    \hline
    ~&DETR&Mask&Faster&Libra&Retina\\
    \hline
    2m&83.3&83.3&0.0&33.3&0.0\\
    \hline
    3m&0.0&60.0&20.0&70.0&0.0\\
    \hline
    4m&0.0&0.0&14.3&0.0&0.0\\
    \hline
    5m&0.0&0.0&0.0&42.1&0.0\\
    \hline
    6m&0.0&0.0&0.0&5.3&0.0\\
    \hline
    
\end{tabular}
\end{table}

\begin{table} 
	\centering
 
    \setlength{\belowcaptionskip}{5pt}
    \caption{Experimental results of physical transfer attacks (Line+Ellipse).}
    \label{Table15}
	\begin{tabular}{cccccc}

    \hline
    ~&DETR&Mask&Faster&Libra&Retina\\
    \hline
    2m&32.1&60.7&32.1&14.3&0.0\\
    \hline
    3m&23.1&38.5&38.5&23.1&0.0\\
    \hline
    4m&2.7&32.4&8.1&78.4&35.1\\
    \hline
    5m&0.0&8.2&2.0&42.9&2.0\\
    \hline
    6m&0.0&31.0&3.4&10.3&13.8\\
    \hline
    
\end{tabular}
\end{table}

\begin{table} 
	\centering
 
    \setlength{\belowcaptionskip}{5pt}
    \caption{Experimental results of physical transfer attacks (Triangle+Ellipse).}
    \label{Table16}
	\begin{tabular}{cccccc}

    \hline
    ~&DETR&Mask&Faster&Libra&Retina\\
    \hline
    2m&33.3&11.1&33.3&11.1&55.6\\
    \hline
    3m&29.2&0.0&4.2&12.5&79.2\\
    \hline
    4m&18.8&31.3&21.9&12.5&46.9\\
    \hline
    5m&24.2&27.3&27.3&39.4&78.8\\
    \hline
    6m&25.0&45.0&15.0&10.0&90.0\\
    \hline
    
\end{tabular}
\end{table}

\begin{table} 
	\centering
 
    \setlength{\belowcaptionskip}{5pt}
    \caption{Experimental results of physical transfer attacks (Line+Triangle+Ellipse).}
    \label{Table17}
	\begin{tabular}{cccccc}

    \hline
    ~&DETR&Mask&Faster&Libra&Retina\\
    \hline
    2m&50.0&25.0&75.0&25.0&50.0\\
    \hline
    3m&0.0&0.0&0.0&42.9&28.6\\
    \hline
    4m&0.0&0.0&5.6&22.2&5.6\\
    \hline
    5m&0.0&0.0&16.7&33.3&16.7\\
    \hline
    6m&13.3&40.0&13.3&0.0&100.0\\
    \hline
    
\end{tabular}
\end{table}

\subsection{The adversarial effect under different pedestrian posture}

\begin{figure}
\centering
\includegraphics[width=1\columnwidth]{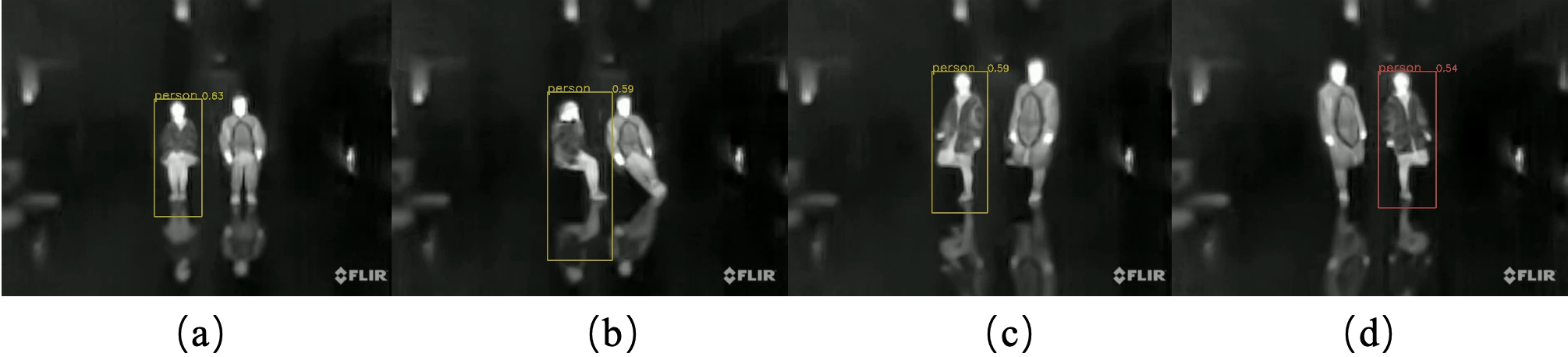} 
\caption{The adversarial effect of AdvIG under different pedestrian posture. There are different sitting and standing positions.}
\vspace{-0.3cm}
\label{figure8}
\end{figure}

We conduct additional experiments to investigate the adversarial effects of AdvIG across different pedestrian postures, as illustrated in Figure \ref{figure8}.  This includes various sitting postures (see Figures \ref{figure8} (a) and (b)) and standing postures (see Figures \ref{figure8} (c) and (d)).  It is evident that AdvIG exhibits adversarial effects across different pedestrian postures.  Furthermore, AdvIG demonstrates effective adversarial attacks even with a certain degree of angular shift.  Upon statistical analysis of these attack experiments, we achieve an overall attack success rate of 73.7\%.

\subsection{Defense of AdvIG}

In addition to investigating the adversarial effects of AdvIG, we also conduct adversarial defense strategies against it. In this experiment, we employ Adversarial Training (AT) \cite{ref14} and DW \cite{ref46} as defense mechanisms. For AT, we utilize randomly generated lines, triangles, and ellipses as perturbations to generate adversarial samples. Subsequently, we fine-tune the Yolo v3 model with a 3:1 ratio of adversarial samples to clean samples. The resulting Yolo v3 model achieves an average accuracy of 90.0\%. As for DW, it encompasses two settings: blind image inpainting and non-blind image inpainting. In blind image inpainting, the defender lacks the position information of the perturbation and must inpaint the image after detecting the perturbation. Conversely, in non-blind image inpainting, the defender can obtain the perturbation's location information and directly repair the perturbation. In this experiment, we opt for non-blind image inpainting as the method for adversarial defense.

We summarize the experimental results in Table \ref{Table18}, leading to the following conclusions: 
(1) Both AT and DW demonstrate effective defense against AdvIG, resulting in a significant decrease in the attack success rate for line, triangle, and ellipse attacks.
(2) Compared to DW, AT exhibits a more effective defense against AdvIG.
(3) Despite the effectiveness of AT and DW in defending against AdvIG, they are unable to completely neutralize its adversarial capabilities.

\begin{table} 
	\centering
 
    \setlength{\belowcaptionskip}{5pt}
    \caption{Defense against AdvIG.}
    \label{Table18}
	\begin{tabular}{ccccc}

    \hline
    ~&~&No defense&AT&DW\\
    \hline
    \multirow{2}*{Line}&ASR&93.1&21.1&33.7\\
    \cmidrule(r){2-5}
    ~&Query&71.7&423.1&382.9\\
    \hline
    \multirow{2}*{Triangle}&ASR&86.8&24.0&41.0\\
    \cmidrule(r){2-5}
    ~&Query&113.1&419.1&304.8\\
    \hline
    \multirow{2}*{Ellipse}&ASR&100.0&43.4&61.7\\
    \cmidrule(r){2-5}
    ~&Query&2.6&293.3&277.5\\
    \hline

\end{tabular}
\end{table}

\section{Conclusion}

In this study, we introduce AdvIG, a straightforward and efficient black-box infrared physical adversarial attack method. AdvIG leverages various geometries to execute adversarial attacks against infrared pedestrian detectors. We evaluate AdvIG based on its effectiveness, stealthiness, and robustness. 

In digital attack experiments, AdvIG consistently achieves an attack success rate of at least 85\%, with an average query count not exceeding 150. Physical attack experiments further demonstrate AdvIG's effectiveness, achieving attack success rates of over 60\%. These results underscore the effectiveness and efficiency of AdvIG.To highlight AdvIG's stealthiness, we compare digital and physical samples generated by AdvIG and baseline methods, confirming its low visibility. Moreover, in robustness tests against advanced object detectors, AdvIG maintains an average attack success rate of no less than 80\%, demonstrating its robustness. Notably, AdvIG outperforms baseline methods in both digital and physical environments.

Given its simplicity, efficiency, stealthiness, and robustness, we advocate for increased attention to AdvIG. Our findings pave the way for future infrared physical attacks, emphasizing the deployment of perturbations inside target objects for enhanced stealthiness. Additionally, we aim to explore multi-view physical attack techniques for black-box scenarios and develop comprehensive defense strategies against infrared physical attacks, representing key areas for future research.





 \bibliographystyle{elsarticle-harv} 
 \bibliography{main}





\end{document}